\documentclass[journal]{IEEEtran}

\IEEEoverridecommandlockouts 

\UseRawInputEncoding
\usepackage{float}
\usepackage{multirow}
\usepackage{graphicx}
\usepackage{placeins}
\usepackage[caption=false]{subfig}
\usepackage{caption}
\usepackage{lscape}
\usepackage{array}

\newcolumntype{P}[1]{>{\centering\arraybackslash}p{#1}}
\usepackage{color}
\usepackage{fancyhdr}
\usepackage{amsmath,amsfonts,amssymb,amsthm}
\usepackage[linesnumbered,ruled,vlined]{algorithm2e}

\usepackage{commath}
\usepackage{mathtools}
\usepackage{enumitem}
\usepackage{pmat}
\usepackage{rotating}
\usepackage{textcomp}
\usepackage{balance}
\usepackage{wrapfig}
\usepackage{hyperref}
\usepackage{booktabs}

\usepackage{xspace} 
\usepackage{microtype}
\usepackage{flushend}
\usepackage{tabularx}
\usepackage[table]{xcolor}
\usepackage{comment}
\usepackage{authblk}
\title{\LARGE \bf Defending Against Gradient Inversion Attacks for Biomedical Images via Learnable Data Perturbation}

\makeatletter
\def\footnoterule{\kern-3\p@
  \hrule \@width 2in \kern 2.6\p@} 
\makeatother

\author{Shiyi Jiang$^{1}$, Farshad Firouzi$^{1,2}$, and Krishnendu Chakrabarty$^{2}$
\footrule
\thanks{$^{1}$Department of Electrical and Computer Engineering, Duke University, Durham, NC 27708 USA. $^{2}$School of Electrical, Computer and Energy Engineering, Arizona State University, Tempe, AZ 85281 USA.}
}

\begin{document}
\maketitle
\begin{abstract}
	The increasing need for sharing healthcare data and collaborating on clinical research has raised privacy concerns. Health information leakage due to malicious attacks can lead to serious problems such as misdiagnoses and patient identification issues. Privacy-preserving machine learning (PPML) and privacy-enhancing technologies, particularly federated learning (FL), have emerged in recent years as innovative solutions to balance privacy protection with data utility; however, they also suffer from inherent privacy vulnerabilities. Gradient inversion attacks constitute major threats to data sharing in federated learning. Researchers have proposed many defenses against gradient inversion attacks. However, current defense methods for healthcare data lack generalizability, i.e., existing solutions may not be applicable to data from a broader range of populations. In addition, most existing defense methods are tested using non-healthcare data, which raises concerns about their applicability to real-world healthcare systems. In this study, we present a defense against gradient inversion attacks in federated learning. We achieve this using latent data perturbation and minimax optimization, utilizing both general and medical image datasets. Our method is compared to two baselines, and the results show that our approach can outperform the baselines with a reduction of 12.5\% in the attacker's accuracy in classifying reconstructed images. The proposed method also yields an increase of over 12.4\% in Mean Squared Error (MSE) between the original and reconstructed images at the same level of model utility of around 90\% client classification accuracy. The results suggest the potential of a generalizable defense for healthcare data.
\end{abstract}


\section{Introduction} \label{sec:intro}
Advances in Artificial Intelligence and Machine Learning (AI/ML) in recent decades have revolutionized various fields, including computer vision and natural language processing \cite{Dargan2019ASO, Firouzi2020IIoT}. Moreover, it has been increasingly applied in healthcare applications, such as medical image diagnosis, gene-disease association detection, and text/voice-based patient care management \cite{Ching2017OpportunitiesAO, Xu2021ChatbotFH}. Recently, there has been an increasing trend of utilizing stored hospital data for clinical research using AI, driven by the ubiquitous digitization of patient health records \cite{Zhang2022ShiftingML}. However, existing hospital data storage systems are vulnerable to adversarial attacks \cite{Seh2020HealthcareDB}. It has been reported that there has been an upsurge in the incidents of data breaches and ransomware attacks on hospitals \cite{Zarour2021EnsuringDI, Neprash2022TrendsIR}. The use of AI technology trained on a large quantity of patient data also poses privacy concerns \cite{Murdoch2021PrivacyAA}. Unauthorized or malicious modifications of patient data by attackers can lead to severe consequences, including identification threats, fatal misdiagnosis, and significant financial losses \cite{Finlayson2019AdversarialAO, Argaw2020CybersecurityOH}.

Federated Learning (FL) offers significant potential for facilitating data sharing between institutions while addressing privacy concerns \cite{Rieke2020TheFO}. FL is a decentralized learning framework that involves a central cloud server coordinating with multiple local clients, such as clinical institutions. FL allows local clients to achieve improved model performance without directly sharing privacy-sensitive data with others, mitigating potential privacy risks \cite{Rieke2020TheFO}. It employs an iterative approach in which clients train their ML model locally, transmit model weights/gradients to the central server, and receive updated model parameters to continue local training \cite{Li2019ASO}. However, recent research has highlighted that FL remains vulnerable to malicious attacks \cite{Li2019ASO}. Such attacks can be performed in either black-box (the attacker has access to the output of a model) or white-box (the attacker has access to the entire model architecture and corresponding parameters) setting and may take place during model training or inference stage \cite{Usynin2021AdversarialIA, Liu2022ThreatsAA}. Attackers can also participate as clients or reside within the central server. For instance, He et al. \cite{He2021AttackingAP} proposed white-box attacks on the edge-cloud collaborative system, assuming the cloud to be untrusted. Yang et al. \cite{Yang2024PracticalFI} introduced black-box feature inference attacks by malicious clients in vertical FL.

There exist many defense techniques to mitigate different attacks. Boulemtafes et al. \cite{Boulemtafes2020ARO} reviewed recent literature on privacy-preserving ML techniques, such as encryption, data perturbation, data projection, partial sharing, and model splitting. They also identified future directions, including the combination of multiple existing methods and the integration of blockchain technology. Nonetheless, most existing defense methods are evaluated using non-healthcare data and thus pose several concerns when applied to the healthcare domain: 1) Medical image data contains biological context, which is domain-specific and hence differs from natural image datasets like CIFAR-10 and ImageNet \cite{Ma2019UnderstandingAA, Kaviani2022AdversarialAA}. 2) Deep learning models developed based on non-healthcare data are usually over-parameterized when applied to medical data \cite{Ma2019UnderstandingAA}. Consequently, they can achieve globally optimal solutions \cite{AllenZhu2018ACT}. This property enables the attackers to reconstruct the private data more easily by exploiting the directions of gradient updates. In addition, existing defense techniques for medical data are limited and primarily rely on generated adversarial examples by leveraging auxiliary datasets. These examples are less effective compared to those generated using natural images due to the scarcity of publicly available medical datasets \cite{Kaviani2022AdversarialAA, Puttagunta2023AdversarialEA}.

In this paper, we focus on defending against gradient inversion attacks in FL, where an attacker by accessing model weights or gradient information, seeks to reconstruct the privacy-sensitive data of the participating clients \cite{Dibbo2023SoKMI, Fang2024PrivacyLO}. To overcome the challenges outlined above, we propose a white-box defense technique that performs latent data perturbation and conducts minimax optimization utilizing the decoder module within the local client. We evaluate our proposed approach on both natural and medical images to ensure its generalizability and effectiveness, particularly in the context of medical data. The key contributions of this paper are summarized as follows.
\begin{enumerate}
    \item We propose a novel defense against gradient inversion attacks in FL on medical image data without the need for auxiliary datasets.
    \item We utilize data perturbation in combination with minimax optimization during client model training, effectively minimizing the risk of information leakage through model gradients.
    \item For the first time, we demonstrate the proposed method on both natural and medical image datasets, showing that our method preserves generalizability and applies to medical data.
    \item The proposed method outperforms two state-of-the-art baselines, achieving a 12.5\% reduction in the attacker's accuracy in classifying reconstructed images, along with a minimum 12.4\% increase in MSE between the original and reconstructed images, while preserving the same level of client model utility.
\end{enumerate}

The remainder of the paper is organized as follows. We provide background information on attacks and defenses on both non-healthcare and healthcare data in Section \ref{sec:related}. In Section \ref{sec:method}, we introduce the proposed defense against gradient inversion attacks in FL. We analyze and discuss the results in Section \ref{sec:results}. Finally, we conclude the paper in Section \ref{sec:conclusion}.
\section{Related Prior Work} \label{sec:related}
\subsection{Adversarial Attacks}
There exist various types of adversarial attacks on machine learning models. For instance, membership inference attacks aim to infer if the given sample exists in the training data; property inference attacks attempt to obtain specific feature information (i.e., gender and age) from private data; data poisoning attacks perturb the input data to degrade the performance of the ML model \cite{Liu2022ThreatsAA}. Nasr et al. \cite{Nasr2018ComprehensivePA} showed that gradient information can be easily utilized under white-box membership inference attacks. Melis et al. \cite{Melis2018ExploitingUF} conducted membership and property inference attacks based on gradient information to demonstrate the existence of privacy concerns in FL. Wang et al. \cite{Wang2019EavesdropTC} proposed attacks that can infer the existence and number of the desired training labels, suggesting the vulnerability of the FL. Wang et al. \cite{Wang2023PoisoningAssistedPI} developed a poisoning-assisted property inference attack that alters the training labels to perturb the global model in FL.

In addition to the attacks mentioned above, model inversion attacks are another category of attacks that cause high privacy risks. Model inversion attacks reconstruct private data from the obtained model update information. Zhang et al. \cite{Zhang2019TheSR} suggested a GAN-based model inversion attack utilizing public auxiliary data that can reconstruct facial images from deep neural networks at a high success rate. Chen et al. \cite{Chen2020KnowledgeEnrichedDM} presented a GAN-based model-inversion attack for distributional recovery utilizing public data and soft labels. An et al. \cite{An.Mirror.NDSS.2022} developed a StyleGAN-based model inversion attack utilizing auxiliary public data. They declared that the reconstructed images achieved state-of-the-art fidelity. Nguyen et al. \cite{Nguyen2023ReThinkingMI} argued their proposed model inversion attack achieved state-of-the-art with optimized objective function and reduced model overfitting.

Besides utilizing GANs and auxiliary public datasets \cite{Fang2024PrivacyLO}, using gradient information to recover private data (gradient inversion attacks) is more prevalent in FL. Zhu et al. \cite{Zhu2019DeepLF} developed deep leakage from gradients (DLG) that reconstructs the victim's private data by emulating victim model gradients. Geiping et al. \cite{Geiping2020InvertingG} proposed an attack termed inverting gradients that reconstructs victim images by minimizing the cosine distance between gradients from the original and reconstructed images plus the total variation of the original images. Wei et al. \cite{Wei2020AFF} investigated the effect of different FL settings on gradient leakage attacks and the impact of different hyperparameters of the attacks on FL. Yin et al. \cite{Yin2021SeeTG} presented GradInversion, which reconstructs a larger batch of data and labels from clients in FL than existing attacks with high fidelity. Wu et al. \cite{Wu2022LearningTI} proposed adaptive attacks to break through existing defense techniques against gradient inversion attacks. Geng et al. \cite{Geng2023ImprovedGI} presented gradient inversion attacks to infer training labels and reconstruct the images in FL. Li et al. \cite{Li2022AuditingPD} proposed a generative gradient leakage attack utilizing both gradient and auxiliary information to existing defense methods, showing the capability of the information leakage with existing defense techniques in FL.

\subsection{Defenses Against Adversarial Attacks}
Many defense algorithms utilize data perturbation to prevent privacy leakage. For instance, Jia and Gong \cite{Jia2018AttriGuardAP} proposed AttriGuard to defend against attribute inference attacks by adding learned noise. Mireshghallah et al. \cite{Mireshghallah2019ShredderLN} proposed to add learned noise to the edge model before sending data to the untrusted cloud server to defend against potential malicious attacks during the inference stage. Sun et al. \cite{Sun2020SoteriaPD} suggested Soteria which learns perturbed data representations that are dissimilar to original input data to defend against model inversion attacks in FL. Mireshghallah et al. \cite{Mireshghallah2021NotAF} suggested learning noise distribution utilizing a pre-trained model and adding learned noise to high-importance features to defend against property inference attacks. Zhu et al. \cite{Zhu2022AFD} built DPFL that adds Laplacian noise to gradients of each layer of the pre-trained DNN model based on layer importance. Shen et al. \cite{Shen2022PerformanceEnhancedFL} created PEDPFL, which improves classification performance at the same privacy level compared to existing DP-based FL algorithms. Cui et al. \cite{Cui2023RecUPFLRU} created RecUP-FL, a method that can defend against user-specified attribute inference attacks and maintain high model utility by adding crafted perturbations to the gradient.

With the rising concerns of DLG attacks in recent years, researchers have proposed defenses that adjust model training objectives to minimize information leakage from model gradients. Roy and Boddeti \cite{Roy2019MitigatingIL} suggested preserving maximal task-specific attribute information while minimizing information of the attribute of interest from the attacker via the measure of entropy. Wang et al. \cite{Wang2020ImprovingRT} suggested a defense against model inversion attacks by minimizing the correlation between the model input and output in terms of mutual information during training. Peng et al. \cite{Peng2022BilateralDO} developed a bilateral dependency optimization strategy that jointly minimizes the correlation between the input and latent space data and maximizes the correlation between the latent space data and the output of a DNN model to defend against model inversion attacks.

Furthermore, there are many other approaches to defend against potential attacks. Hamm \cite{Hamm2016MinimaxFL} suggested a minimax filter to transform the data with an optimal utility-privacy trade-off to defend against inference attacks. Stock et al. \cite{Stock2023LessonsLD} proposed property unlearning to defend against property inference attacks. However, they further noted that the method cannot be generalized to arbitrary property inference attacks with explanations utilizing the explainable AI tool provided. Gong et al. \cite{Gong2023AGD} designed a GAN-based defense method against white-box model inversion attacks utilizing generated fake data learned from local client data and publicly available datasets together with continual learning. Sandeepa et al. \cite{Sandeepa2024RecDefAR} proposed to defend against membership inference and DLG attacks in FL utilizing privacy recommendation systems within local clients. Additionally, Boulemtafes et al. \cite{Boulemtafes2020ARO} summarized existing privacy-preserving ML techniques such as encryption, partial sharing, and model splitting in their survey of recent literature.

There also exist some open challenges that are not addressed by proposed defenses. For instance, Dibbo \cite{Dibbo2023SoKMI} pointed out the lack of generalizability and optimization on the trade-off between privacy and model utility in existing defense methods. Recent surveys \cite{Chen2022FederatedLA, Li2024ThreatsAD} have also pointed out the difficulty of balancing security with communication and computing costs. Furthermore, existing defenses focus merely on horizontal FL rather than other scenarios, such as vertical FL. Additionally, previous work has not developed fairness-aware methods that consider client bias and client benefits versus their corresponding contributions to FL.

\subsection{Attacks and Defenses for Healthcare}
Attacks aimed specifically at healthcare systems have received relatively less attention. Mozaffari-Kermani et al. \cite{Kermani2015SystematicPA} presented a data poisoning attack on multiple medical datasets and proposed a defense method to mitigate information leakage. Newaz et al. \cite{Newaz2020AdversarialAT} utilized multiple adversarial attacks on medical devices to alter the physiological readings of the patients. The results suggested that medical devices are vulnerable to malicious attacks. They further pointed out the lack of existing solutions to defend against the attacks. Rhaman et al. \cite{Rahman2020AdversarialET} developed adversarial attacks on COVID-19 diagnostic deep learning models on medical IoT devices, showing the susceptibility of medical devices to attacks. Ma et al. \cite{Ma2019UnderstandingAA} applied multiple types of existing adversarial attacks to medical imaging datasets and found that medical data is more susceptible to attacks than non-medical data. However, they further claimed that the detection techniques can easily detect the attacks and will not negatively impact the diagnosis results when a physician is involved \cite{Ma2019UnderstandingAA}. Grama et al. \cite{Grama2020RobustAF} showed that data poisoning on healthcare data can be effectively defended with a combination of data perturbation and robust aggregation techniques. Lin et al. \cite{Lin2023PrivacyAwareAC} proposed an attribute-based Secure Access Control Mechanism (SACM) to provide authorization based on users' social data utilizing graph neural networks. Ali et al. \cite{Ali2022FederatedLF} surveyed existing literature on privacy-preserving FL for the Internet of Medical Things (IoMT) and pointed out the need to develop generalizable solutions specifically for smart healthcare systems.
According to Muoka et al. \cite{Muoka2023ACR}, the limited availability of medical data can significantly impair the generalizability of existing defenses, which primarily depend on synthesized adversarial examples from auxiliary datasets \cite{Kaviani2022AdversarialAA}. In addition, Muoka et al. \cite{Muoka2023ACR} raised concerns about the lack of transferability in defending against universal attacks in existing defenses for healthcare data. The above-mentioned drawbacks in the current approaches highlight the need to create more effective solutions for protecting healthcare data.
\section{Proposed Method} \label{sec:method}
\subsection{Problem Setup}
\paragraph*{Threat Model}
The threat model in this work is framed under the federated learning environment, which consists of a central server and some participating clients. The honest but curious central server acts as the attacker and one of the participating clients serves as the victim. During each iteration of communication between the clients and the central server in FL, the server (attacker) receives local model parameters/gradients from the clients. The server (attacker) performs white-box gradient inversion attacks utilizing the victim's model gradients to reconstruct the private dataset of the victim client.

\paragraph*{Defender}
The defender has prior knowledge that the attacker will use model weights/gradients to recover local data. To defend against the gradient inversion attacks by the central server, the victim (defender), possessing both local data and corresponding labels, aims to minimize the amount of input data information encoded within model weights/gradients. The defender leverages learnable data perturbation and minimax optimization with an additional decoder module, detailed in the following sections, to prevent information leakage from the model weights/gradients.


\subsection{Learnable Data Perturbation}

\begin{figure*}[hbt!]
\includegraphics[width=.8\textwidth]{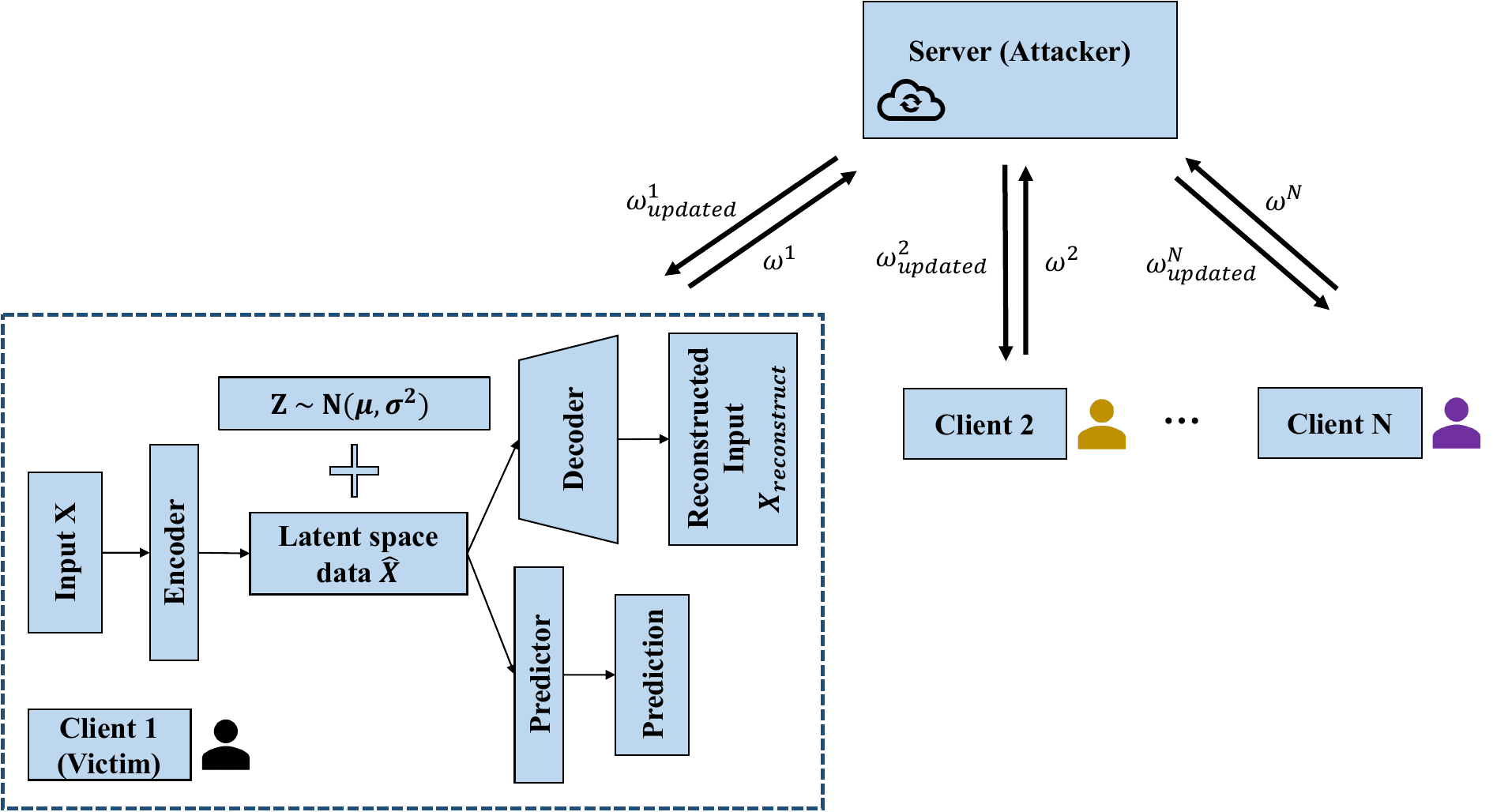}
\centering
\caption{Overall framework of the proposed method for defending against reconstruction attacks.}
\label{fig:framework}
\end{figure*}


We present our proposed method in Fig. \ref{fig:framework}. In the victim's local model, Gaussian noise is injected into latent space data after the encoder layers. In addition to the predictor layers for the conventional classification, we establish an additional branch of decoder layers to locally reconstruct input data. During client model training, we jointly minimize the correlation between the original and reconstructed input data while simultaneously maximizing the classification loss.

We first present the corresponding architectures of the encoder, decoder, and predictor within the defender in Table \ref{tab:arch}. For convolutional and transposed convolutional layers, $c\_out$ denotes the number of output channels, $k$ represents the kernel size, and $s$ indicates the stride. The values and variables in the bracket of the linear layer indicate the input and output dimensions, where $NClass$ represents the number of classes. We use the rectified linear unit (ReLU) activation function after each convolutional layer in the encoder module and after each batch normalization layer in the decoder module. Note that the proposed architectures are adapted from the work of Geiping et al. \cite{Geiping2020InvertingG} and are not designed for a specific task. Additionally, according to \cite{Geiping2020InvertingG}, increases in the width or depth of the model architecture do not decrease the risks of successful data recovery by the attacker.

\begin{table}[hbt!]
\centering
\caption{Details of the defender architecture.}
\label{tab:arch}
\begin{tabular}{c}
\hline
Encoder \\ \hline
Conv2d (c\_out=64, k=3, s=1) \\
Conv2d (c\_out=128, k=3, s=1) \\
Conv2d (c\_out=128, k=3, s=1) \\
Conv2d (c\_out=256, k=3, s=1) \\
Conv2d (c\_out=256, k=3, s=1) \\
Conv2d (c\_out=256, k=3, s=1) \\
MaxPool2d (k=3) \\
Conv2d (c\_out=256, k=3, s=1) \\
Conv2d (c\_out=256, k=3, s=1) \\
Conv2d (c\_out=256, k=3, s=1) \\ \hline
Decoder \\ \hline
ConvTranspose2d (c\_out=256, k=3, s=1) \\
BatchNorm2d(256) \\
ConvTranspose2d (c\_out=256, k=4, s=1) \\
BatchNorm2d(256) \\
ConvTranspose2d (c\_out=3, k=2 (CIFAR-10) / 4 (BloodMNIST), s=2) \\ \hline
Predictor \\ \hline
MaxPool2d (k=3) \\
Linear (2304, NClass) \\ \hline
\end{tabular}
\end{table}

We denote the defender's private data as $X = \{x_i,\ i = 1, ..., n\}$ with corresponding labels $Y = \{y_i,\ i = 1, ..., n\}$, where $n$ is the number of samples within the defender. After feeding $X$ into the encoder, we obtain transformed data $\hat{X} = \{\hat{x}_i,\ i = 1, ..., n\}$ in the latent space. We sample noise $Z = \{z_i,\ i = 1, ..., n\}$ from the Gaussian distribution $N(\mu, \sigma^2)$ and add $Z$ to $\hat{X}$. The resulting perturbed latent space data is then fed into both the predictor and the decoder simultaneously.

We considered two variations of the noise sampling method, as outlined below:
\paragraph{\textbf{Noise sampled from distributions with fixed parameters}}
In this scenario, we use noise sampled from Gaussian distributions with the same parameter values $\mu$ and $\sigma$ defined by the defender, i.e., $Z = \{z_i \sim N(\mu, \sigma^2),\ i = 1, ..., n\}$. Note that this approach may be utilized only when the defender has prior knowledge of the impact of noise levels on its dataset. The training procedure follows conventional federated learning, with iterative local model training on the client side, followed by the upload and download of the server and the client.

\paragraph{\textbf{Noise sampled from learned distributions}}
In this scenario, we require a pre-training procedure to learn the optimal noise distributions to be applied to each element of the latent space data before the client is involved in the federated learning. Initially, we sample noise from Gaussian distributions initialized with the same parameter values defined by the defender. We then perform local client training to learn the optimal parameters for these Gaussian distributions, adjusting the noise added to each element, i.e., $Z = \{z_{ij} \sim N(\mu_j, \sigma_j^2),\ i = 1, ..., n,\ j = 1, ..., m\}$, where $m$ is the number of elements within each latent space data $\hat{x}_i$. Once the optimal noise distributions are learned, we conduct federated learning with the client model weights initialized randomly, while retaining the learned noise distributions for use throughout training.

We use the Pearson correlation coefficient $r$ \cite{Benesty2009PearsonCC}, as shown in Equation (\ref{eq:1}), to quantify the correlations between the input $x$ and the reconstructed data $x_{reconstruct}$, where $\mu_x$ and $\mu_{x_{reconstruct}}$ represent the means of $x$ and $x_{reconstruct}$, and $x_i$ and $x_{reconstruct, i}$ denote single elements of $x$ and $x_{reconstruct}$, respectively. Prior research has demonstrated that the Pearson correlation coefficient effectively reveals statistical similarity between the original and synthesized medical datasets \cite{Sun2023GeneratingSP}.
\begin{equation}\label{eq:1}
    r = \frac{\sum_i (x_i - \mu_x)(x_{reconstruct, i} - \mu_{x_{reconstruct}})}{\sqrt{\sum_i (x_i - \mu_x)^2 \cdot \sum_i (x_{reconstruct, i} - \mu_{x_{reconstruct}})^2}}
\end{equation}

During model training, we first update the decoder module that attempts to maximize the correlation between the original image $x$ and the reconstructed image $x_{reconstruct}$ using Equation (\ref{eq:2}) as the loss function.
\begin{equation}\label{eq:2}
    l_{decoder} = 1 - |r_{x, x_{reconstruct}}|
\end{equation}
Next, the predictor module is updated using Equation (\ref{eq:3}) as the loss function to maximize the classification accuracy while minimizing the correlation between $x$ and $x_{reconstruct}$. In this equation, $l_{CE} (\cdot)$ represents the cross-entropy loss \cite{Ho2020TheRC}, $y$ denotes the ground truth, $\hat{y}$ indicates the output from the predictor module, and $\alpha$ serves as a scalar coefficient.
\begin{equation}\label{eq:3}
    l_{predictor} = l_{CE} (y, \hat{y}) + \alpha \cdot |r_{x, x_{reconstruct}}|
\end{equation}

To summarize, we provide the pseudocode of the local pre-training procedure of the defender with noise sampled from learnable distributions in Algorithm \ref{algo:1}, where $\lambda$ is the learning rate. In this approach, only the learned noise hyperparameters will be utilized for federated learning as the next step. Alternatively, for the defender with noise sampled from distributions with fixed parameters, the same procedure is followed during local client training in federated learning, with the key difference being that the noise distribution parameters remain fixed and are not trainable.

\begin{algorithm}[hbt!]
\KwIn{Defender (D), NumEpoch, $\alpha$, $\lambda$}
\KwOut{Trained Defender (D)}
Initialize(D.model)\;
\tcc{Note that D.model contains trainable noise distribution parameters.}
\While{epoch $\leq$ NumEpoch}{
    \For{$x$, $y$ in D.loader}{
        $x_{reconstruct},\ \hat{y} \gets D.model(x)$\;
        $l_{decoder} \gets 1 - |r_{x, x_{reconstruct}}|$\;
        $\omega_{D.model} \gets \omega_{D.model} - \lambda \cdot \nabla_{D.model}~l_{decoder}$\;
        $l_{predictor} \gets l_{CE}(y, \hat{y}) + \alpha \cdot |r_{x, x_{reconstruct}}|$\;
        $\omega_{D.model} \gets \omega_{D.model} - \lambda \cdot \nabla_{D.model}~l_{predictor}$\;
    }
}
\caption{Pre-training procedure of the defender.}
\label{algo:1}
\end{algorithm}
\section{Results} \label{sec:results}
\subsection{Experimental Setting}
\subsubsection{Datasets}
We used the following two datasets to evaluate the proposed method.

\paragraph*{\textbf{CIFAR-10}}
The CIFAR-10 dataset is a collection of 60,000 $32 \times 32$ RGB images containing 10 classes of different objects including animals and vehicles \cite{Krizhevsky2009LearningML}. Each class contains 6000 images. The training set contains 50,000 images and the test set contains 10,000 images.

\paragraph*{\textbf{BloodMNIST}}
The BloodMNIST dataset was introduced recently for biomedical image analysis \cite{medmnistv2}. The dataset is composed of 17,092 $28 \times 28$ RGB images. Each image captures an individual peripheral blood cell sourced from the Hospital Clinic of Barcelona. There are eight different classes of cells in total.

\subsubsection{Evaluation Metrics}
To evaluate the similarity between the original images and reconstructed images learned by the attacker, we follow metrics used in the attack method \cite{Geiping2020InvertingG} utilized in this study: mean square error (MSE) and peak signal-to-noise ratio (PSNR). In addition, we report client classification accuracy and F1 score as measurements of model utility.

\subsubsection{Baseline Methods}
We compare our method with the following two baseline methods.
\begin{itemize}
    \item Differentially Private Stochastic Gradient Descent (DP-SGD) \cite{Abadi2016DeepLW}: It serves as a benchmark that prevents information leakage by adding noise sampled from the Gaussian distribution with zero mean and user-defined variance to the gradients during model training.
    \item Bilateral Dependency Optimization (BiDO) \cite{Peng2022BilateralDO}: It defends against gradient inversion attacks by maximizing the correlation between latent features and the output and minimizing the correlation between latent features and the input data simultaneously.
\end{itemize}

\subsubsection{Hyperparameter Selection} \label{sec:parameter}
We applied normalization to the datasets and followed their default train/test data division settings. We used a batch size of 128 by convention \cite{He2015DeepRL, medmnistv2} for model training for both datasets.

For image reconstruction, we utilized a recently developed gradient inversion attack presented in \cite{Geiping2020InvertingG}. We adopted the setting of reconstructing images with a batch size of eight using model weight update information for five local epochs. We applied a learning rate of $1 \times 10^{-4}$ for local client model training and $1$ for attacker model training using the CIFAR-10 dataset. We used a learning rate of $1 \times 10^{-3}$ for local client model training and $1 \times 10^{-2}$ for attacker model training using the BloodMNIST dataset.

We present training hyperparameters used in Table \ref{tab:setup} for the evaluation of client model utility after applying different defense methods. Note that except for the DP-SGD baseline using the SGD optimization, we used Adam optimization for the rest of the model training.

\begin{table}[hbt!]
\centering
\caption{Hyperparameters for client model training using CIFAR-10 and BloodMNIST.}
\label{tab:setup}
\begin{tabular}{@{}ccccc@{}}
\hline
\multirow{2}{*}{} & \multicolumn{2}{c}{CIFAR-10} & \multicolumn{2}{c}{BloodMNIST} \\
 & Learning Rate & Epoch & Learning Rate & Epoch \\ \hline
Proposed Method & $1 \times 10^{-4}$ & 400 & $1 \times 10^{-3}$ & 400 \\
DP-SGD & $1 \times 10^{-2}$ & 150 & $1 \times 10^{-1}$ & 200 \\
BiDO & $1 \times 10^{-4}$ & 200 & $1 \times 10^{-3}$ & 200 \\ \hline
\end{tabular}
\end{table}

For the proposed defense method with a fixed noise distribution, we used the following values, $1 \times 10^{-4}, 1 \times 10^{-3}, 1 \times 10^{-2}, 1 \times 10^{-1}, 1$, and $10$, as the mean of the noise distribution with a fixed standard deviation of 0.1. We initialized the noise distribution with a mean of 1 and a standard deviation of 0.1 for the proposed method with learnable noise distribution. We set the regularization hyperparameter $\alpha = 1$ for model training.

We used values $1 \times 10^{-4}, 1 \times 10^{-3}, 1 \times 10^{-2}, 1 \times 10^{-1}, 1$, and $10$ as the standard deviation of the noise distribution for DP-SGD baseline. We tested hyperparameters of the BiDO baseline $(\lambda_x, \lambda_y) \in [(1, 5),\ (0.5, 5),\ (0.1, 2),\ (0.1, 3),\ (0.1, 4),\ (0.1, 5)]$ with the CIFAR-10 dataset and $(\lambda_x, \lambda_y) \in [(2, 10),\ (1, 10),\ (0.5, 10),\ (0.1, 3),\ (0.5, 20),\ (0.1, 5)]$ with the BloodMNIST dataset.

All experiments were executed using the Scikit-learn \cite{scikit-learn} and the PyTorch package \cite{NEURIPS2019_9015}.

\subsubsection{Platform} All experiments were executed using two NVIDIA GeForce RTX 2080Ti each with 11 GB RAM and two NVIDIA TITAN V each with 12 GB RAM.

\subsection{Comparison with Baseline Methods}

\begin{figure}[htb!]
\subfloat[Original]{%
  \centering
  \includegraphics[width=\columnwidth]{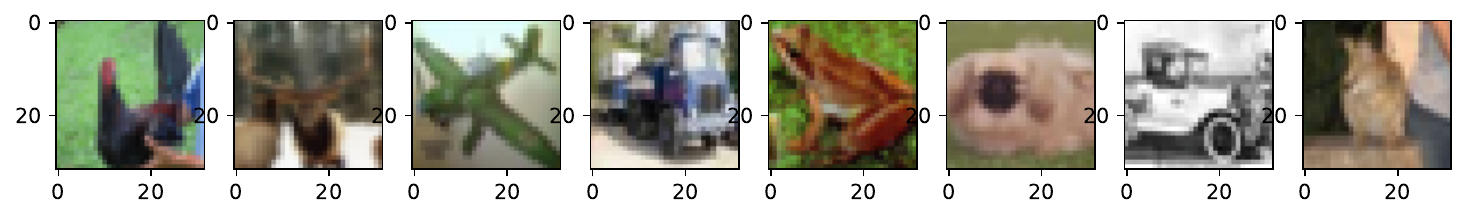}%
}

\subfloat[No defense method]{%
  \centering
  \includegraphics[width=\columnwidth, trim={0 0 5.8cm 0.7cm}, clip]{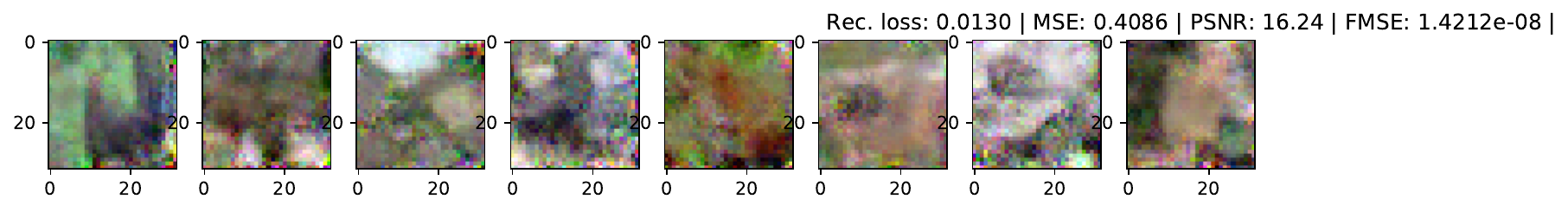}%
}

\subfloat[DP-SGD, $\sigma=1 \times 10^{-4}$]{%
  \centering
  \includegraphics[width=\columnwidth, trim={0 0 5.8cm 0.7cm}, clip]{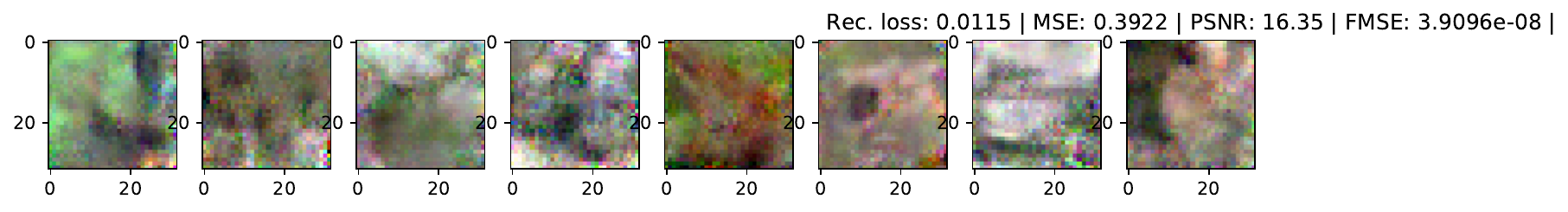}%
}

\subfloat[BiDO, $(\lambda_x, \lambda_y) = (1, 10)$]{%
  \centering
  \includegraphics[width=\columnwidth, trim={0 0 5.8cm 0.7cm}, clip]{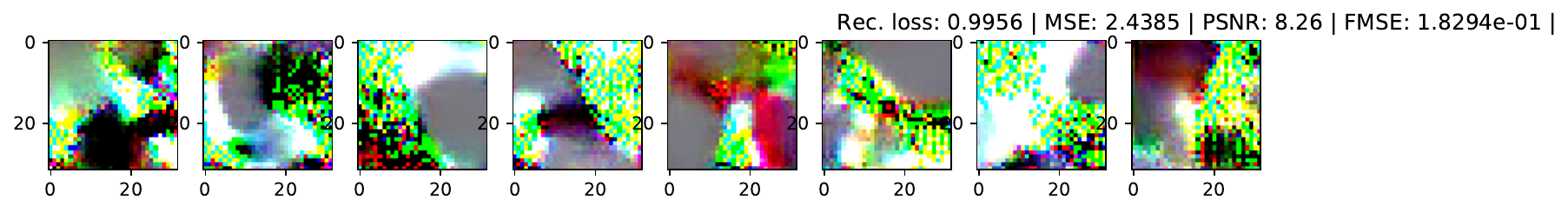}%
}

\subfloat[Proposed method, $\mu = 1 \times 10^{-3}$]{%
  \centering
  \includegraphics[width=\columnwidth, trim={0 0 5.8cm 0.7cm}, clip]{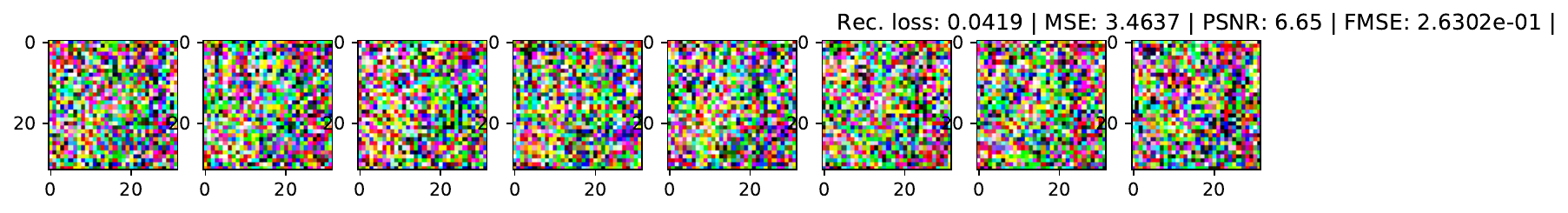}%
}
\caption{Comparisons between original sample images from CIFAR-10 and reconstructed images using different defense methods.}
\label{img:cifar-10}
\end{figure}

\begin{figure}[htb!]
\subfloat[Original]{%
  \centering
  \includegraphics[width=\columnwidth]{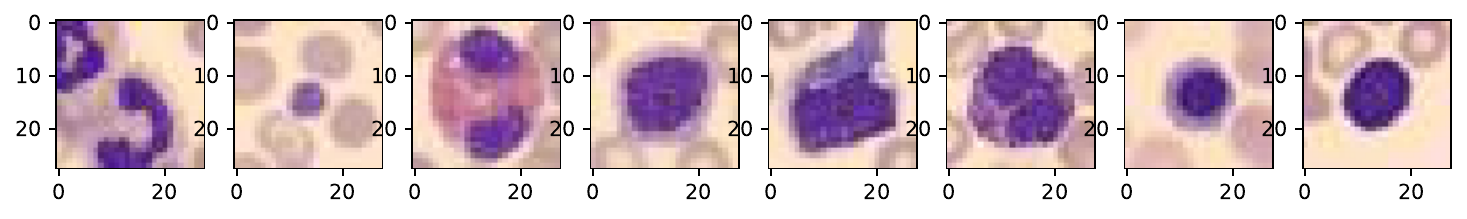}%
}

\subfloat[No defense method]{%
  \centering
  \includegraphics[width=\columnwidth, trim={0 0 5.8cm 0.7cm}, clip]{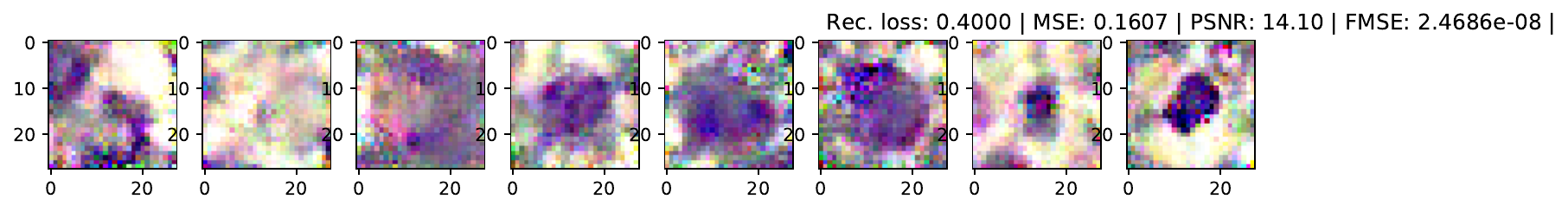}%
}

\subfloat[DP-SGD, $\sigma=1 \times 10^{-4}$]{%
  \centering
  \includegraphics[width=\columnwidth, trim={0 0 5.8cm 0.7cm}, clip]{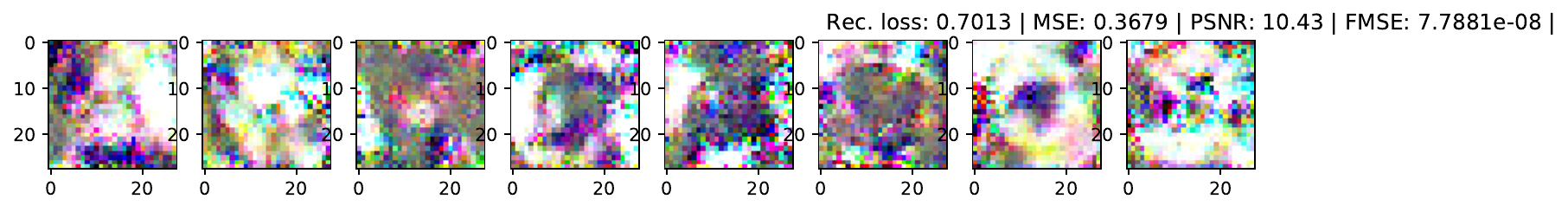}%
}

\subfloat[BiDO, $(\lambda_x, \lambda_y) = (0.1, 4)$]{%
  \centering
  \includegraphics[width=\columnwidth, trim={0 0 5.8cm 0.7cm}, clip]{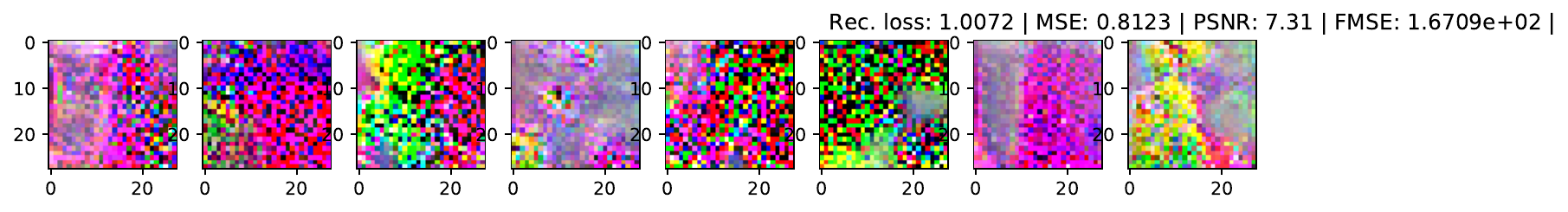}%
}

\subfloat[Proposed method, $\mu = 1 \times 10^{-4}$]{%
  \centering
  \includegraphics[width=\columnwidth, trim={0 0 5.8cm 0.7cm}, clip]{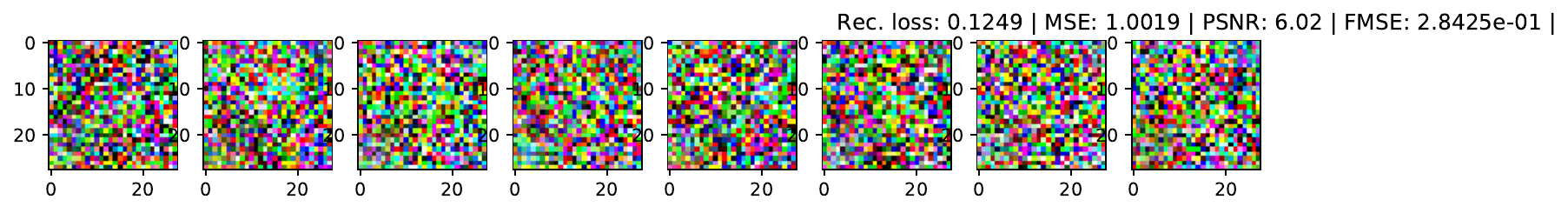}%
}
\caption{Comparisons between original sample images from BloodMNIST and reconstructed images using different defense methods.}
\label{img:bloodmnist}
\end{figure}

We present original sample images and their corresponding reconstructed images with different defense methods in Fig. \ref{img:cifar-10} and Fig. \ref{img:bloodmnist} using two datasets. We observe that the attacker can easily reconstruct clear original images without any defense. DP-SGD at a low noise level can also reveal most of the information from the reconstructed images. However, by visual observation, we do not obtain any valuable information from reconstructed images obtained with BiDO and the proposed method.

To comprehensively compare the proposed and baseline methods, we chose ranges of their corresponding hyperparameters for noise generation as described in Section \ref{sec:parameter}. We present the difference between the original and reconstructed images in MSE versus client classification accuracy with different defense methods in Fig. \ref{fig:sweep} to evaluate the robustness of the defense methods in terms of model utility and the successfulness of the attacker. We observe that DP-SGD can achieve high MSE when sacrificing classification accuracy. The proposed method achieves comparable classification accuracy with BiDO, but our method yields higher MSE compared to BiDO. In addition, we notice that the proposed method using adaptive noise shows comparable metrics with the method using fixed noise distributions.

\begin{figure}[H]
\centering
\subfloat[CIFAR-10]{%
  \includegraphics[width=.85\columnwidth]{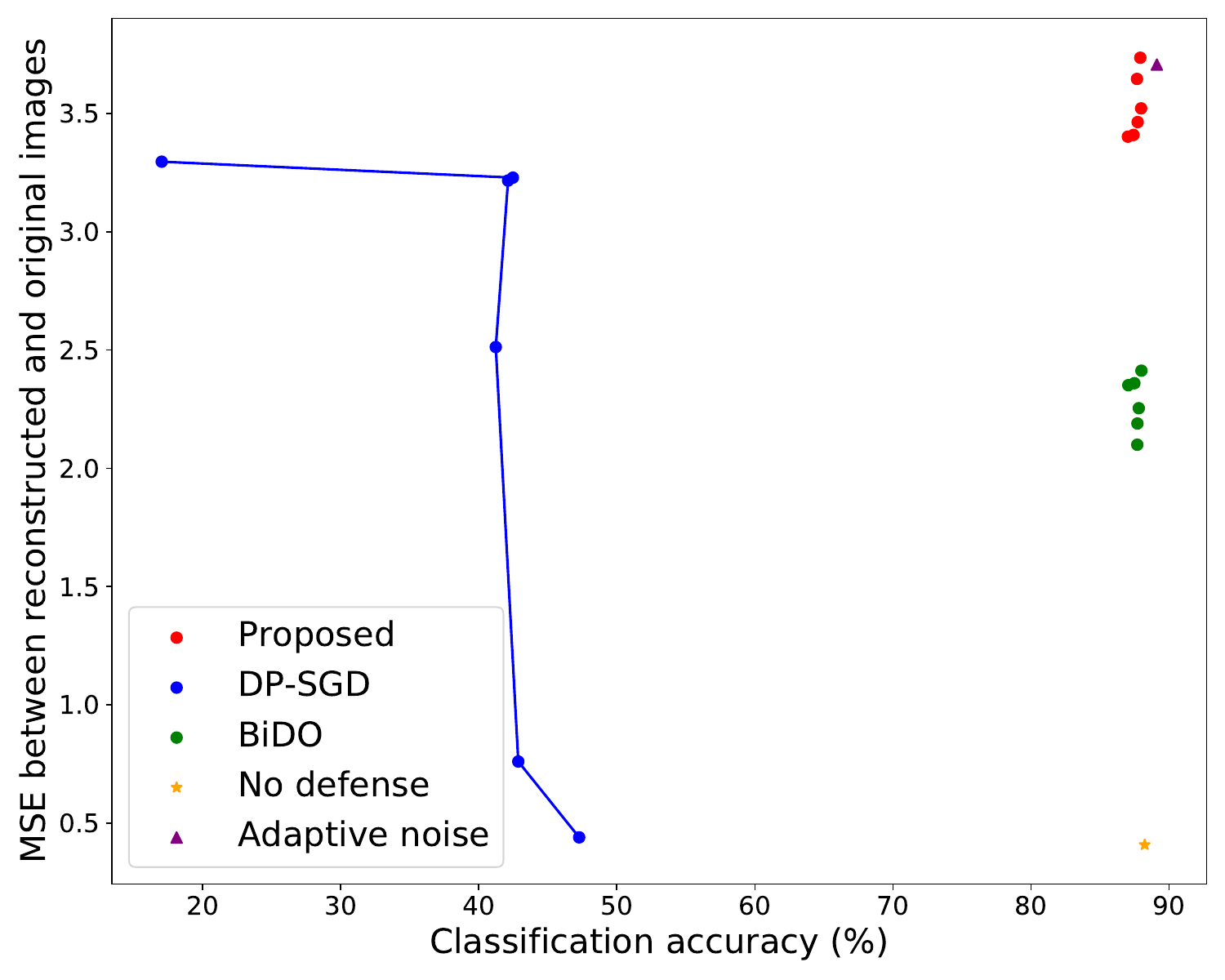}%
}
\end{figure}
\clearpage

\begin{figure}[H]\ContinuedFloat
\centering
\subfloat[BloodMNIST]{%
  \includegraphics[width=.85\columnwidth]{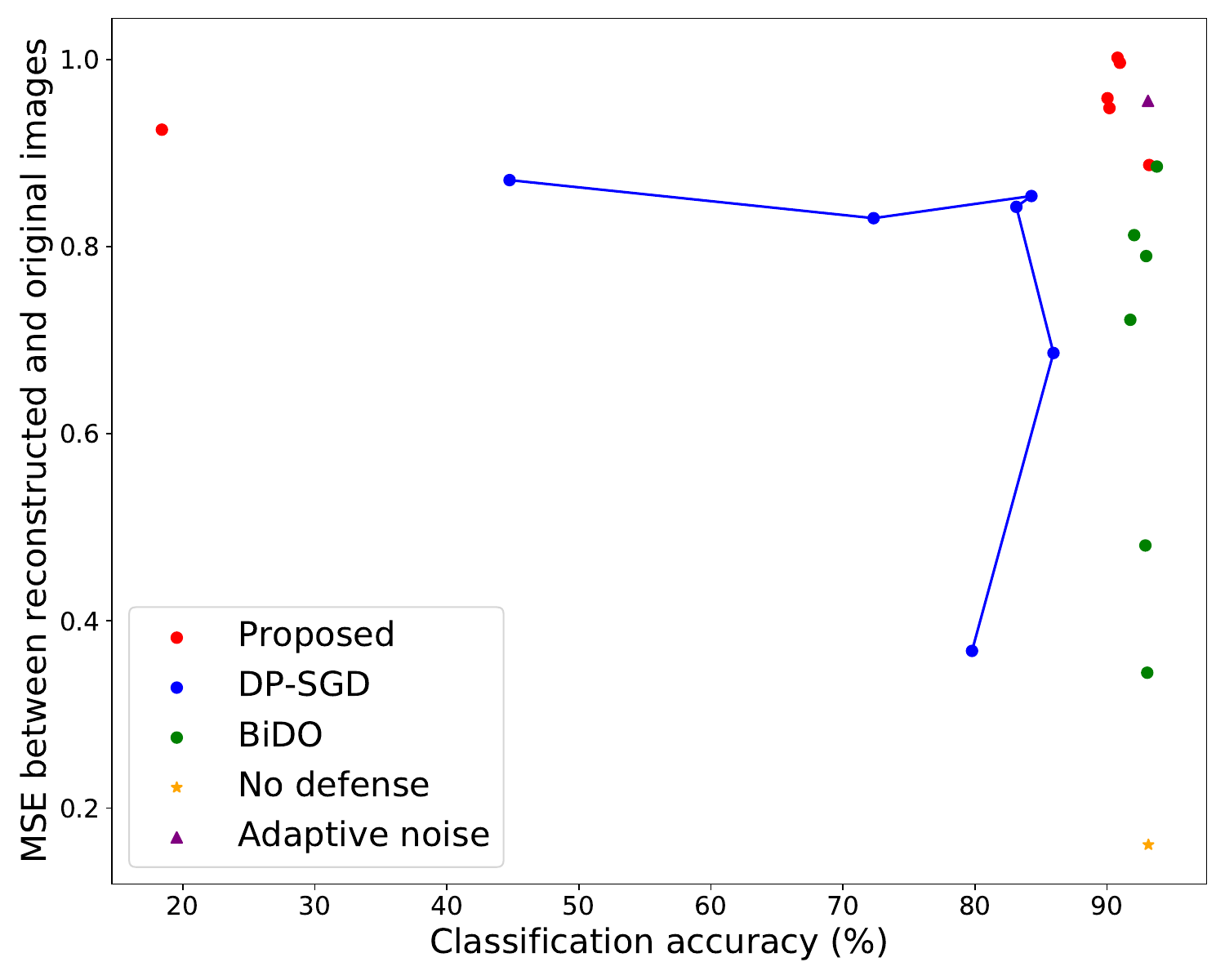}%
}
\caption{Similarity between the original and reconstructed images in terms of MSE versus client classification accuracy.}
\label{fig:sweep}
\end{figure}

In addition to the MSE, we also present the PSNR values for the two datasets using different defense methods shown in Table \ref{tab:proposed}, Table \ref{tab:dpsgd}, and Table \ref{tab:BiDO}. From the tables, we find that the proposed method yields the lowest PSNR compared to the baseline methods while maintaining satisfactory client classification accuracy. To summarize, the proposed method can result in the attacker reconstructing the most dissimilar images to the original images compared to the baseline methods at the same level of client classification accuracy.

\begin{table}[!htbp]
\centering
\resizebox{\columnwidth}{!}{%
\begin{tabular}{@{}ccccccc@{}}
\toprule
Hyperparameter & $1 \times 10^{-4}$ & $1 \times 10^{-3}$ & $1 \times 10^{-2}$ & $1 \times 10^{-1}$ & $1$ & $10$ \\ \midrule
CIFAR-10 & 16.35 & 13.28 & 8.05 & 6.94 & 6.95 & 6.86 \\
BloodMNIST & 10.43 & 7.68 & 6.78 & 6.71 & 6.84 & 6.63 \\ \bottomrule
\end{tabular}%
}
\caption{PSNR (dB) between the original and reconstructed images using DP-SGD \cite{Abadi2016DeepLW} with different noise parameters.}
\label{tab:dpsgd}
\end{table}

\begin{table}[!htbp]
\centering
\resizebox{\columnwidth}{!}{%
\begin{tabular}{@{}ccccccc@{}}
\toprule
Hyperparameter & (2, 10) & (1, 10) & (0.5, 10) & (0.1, 3) & (0.5, 20) & (0.1, 5) \\ \midrule
CIFAR-10 & 9.18 & 8.5 & 8.64 & 8.48 & 9.33 & 8.82 \\ \midrule
Hyperparameter & (1, 5) & (0.5, 5) & (0.1, 2) & (0.1, 3) & (0.1, 4) & (0.1, 5) \\ \midrule
BloodMNIST & 9.29 & 7.73 & 11.17 & 7.26 & 7.31 & 7 \\ \bottomrule
\end{tabular}%
}
\caption{PSNR (dB) between the original and reconstructed images using BiDO \cite{Peng2022BilateralDO} with different hyperparameters.}
\label{tab:BiDO}
\end{table}

\begin{table}[!htbp]
\centering
\resizebox{\columnwidth}{!}{%
\begin{tabular}{@{}ccccccc@{}}
\toprule
Hyperparameter & $1 \times 10^{-4}$ & $1 \times 10^{-3}$ & $1 \times 10^{-2}$ & $1 \times 10^{-1}$ & $1$ & $10$ \\ \midrule
CIFAR-10 & 6.72 & 6.65 & 6.43 & 6.72 & 6.32 & 7.79 \\
BloodMNIST & 6.02 & 6.21 & 6.05 & 6.26 & 6.55 & 6.37 \\ \bottomrule
\end{tabular}%
}
\caption{PSNR (dB) between the original and reconstructed images using the proposed method with different noise parameters.}
\label{tab:proposed}
\end{table}


\subsection{Effect of MSE Difference on the Classification Accuracy of Reconstructed Images}
\begin{figure}[htb!]
\centering
\includegraphics[width=.85\columnwidth]{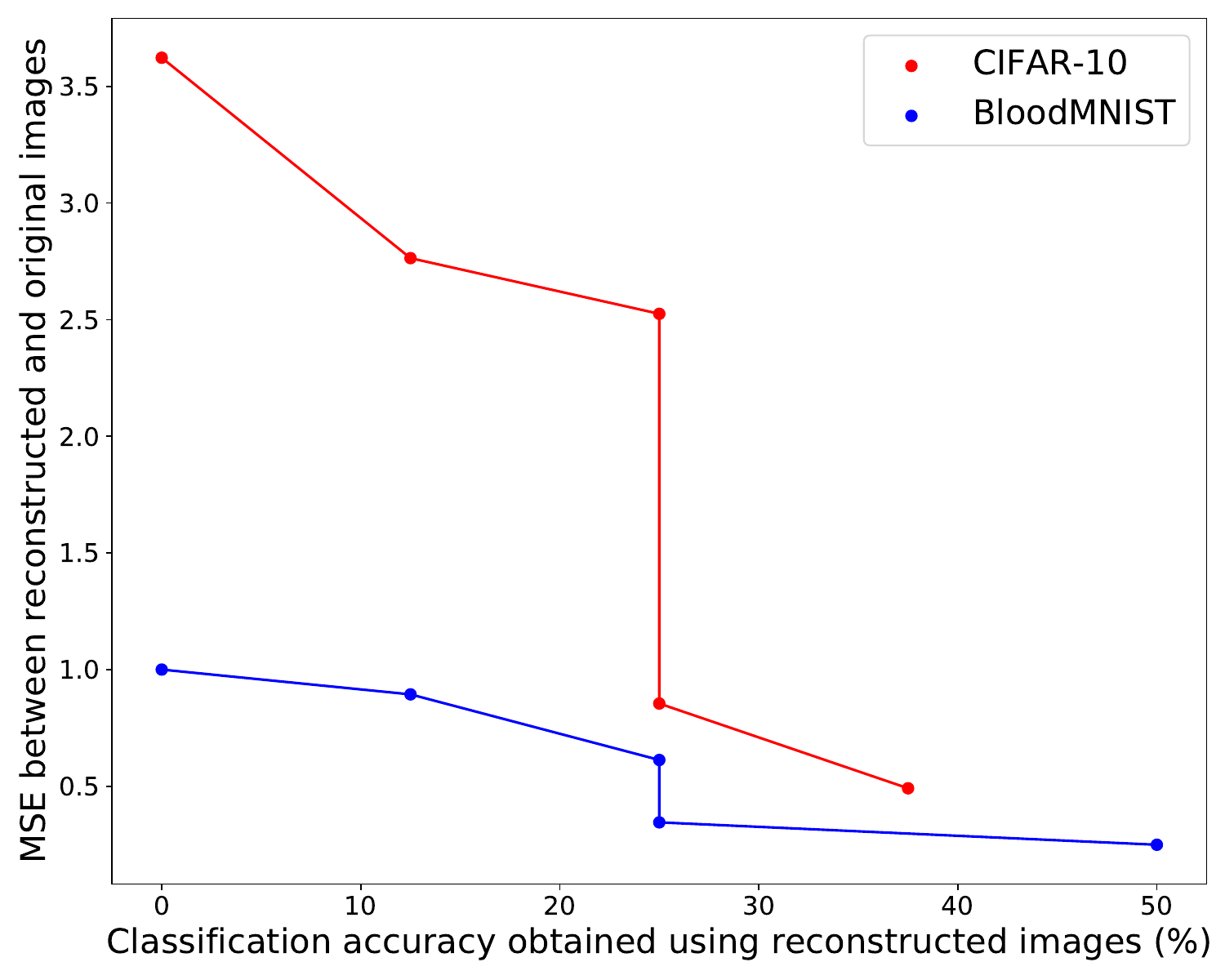}%
\caption{Similarity between the original and reconstructed images in terms of MSE versus client classification accuracy.}
\label{fig:validate}
\end{figure}

We next investigate how the increase in MSE between the original and reconstructed images affects the classification accuracy of the reconstructed images. We selected reconstructed images at different MSE levels and computed the corresponding classification accuracy using the trained classification model without the defense method. We present the results in Fig. \ref{fig:validate}.

We observe that an increase in MSE from 2.8 to 3.6 results in a decrease in accuracy from 12.5\% to 0\% using the CIFAR-10 dataset. Similarly, an increase in MSE from 0.89 to 1.0 leads to a decrease in accuracy down to 0\% using the BloodMNIST dataset. The results further indicate that the proposed method outperforms BiDO and DP-SGD with the increase in MSE leading to significant information perturbation on reconstructed images from the attacker. We thus conclude that the proposed method effectively prevents the attacker from using gradient information while maintaining sufficiently good model utility.

Note that the runtime complexity of the proposed method and the baselines are the same given the same client model. However, the proposed method requires a pre-training step to learn noise distributions, resulting in additional computational cost that depends on the depth, width, and filter sizes of the client model \cite{He2014ConvolutionalNN}. Despite this extra cost, the proposed method shows its effectiveness in preventing attackers from reconstructing private client data. Therefore, we conclude that it outperforms the baselines with acceptable computational cost.

\subsection{Limitations and Future Work}
In this work, we have conducted gradient inversion attacks to demonstrate the effectiveness of the proposed defense method. Additional types of attacks, such as data poisoning attacks and property inference attacks \cite{Lyu2020ThreatsTF}, can be incorporated for a more comprehensive evaluation. Furthermore, research shows that the attacker can learn the defense mechanism of the defender after some training epochs \cite{Wu2022LearningTI, Yang2024ALA}. We will further investigate preventing such reverse engineering techniques in future work. In addition, Huang et al. \cite{Huang2021EvaluatingGI} suggested that combining defense techniques can be more effective in defending against gradient inversion attacks with less model utility loss. We will consider these directions in future work.
\section{Conclusion} \label{sec:conclusion}
Healthcare data exhibits high risks of information leakage and is vulnerable to malicious attacks. Existing defense methods against these attacks are designed based on non-healthcare data, which raises concerns about their effectiveness for healthcare data. In our research, we have developed a defense method against gradient inversion attacks in federated learning. Our method provides a higher mean squared error (MSE) and a lower peak signal-to-noise ratio (PSNR) between the original and reconstructed data from the attacker, while still maintaining a comparable classification accuracy of around 90\% compared to the baseline methods. Furthermore, we have demonstrated the effectiveness of our method using both natural and medical imaging datasets. We will investigate defense methods that enhance the privacy-utility trade-off and prevent reverse engineering by the attacker in future work.


\begin{thebibliography}{10}
\providecommand{\url}[1]{#1}
\csname url@samestyle\endcsname
\providecommand{\newblock}{\relax}
\providecommand{\bibinfo}[2]{#2}
\providecommand{\BIBentrySTDinterwordspacing}{\spaceskip=0pt\relax}
\providecommand{\BIBentryALTinterwordstretchfactor}{4}
\providecommand{\BIBentryALTinterwordspacing}{\spaceskip=\fontdimen2\font plus
\BIBentryALTinterwordstretchfactor\fontdimen3\font minus \fontdimen4\font\relax}
\providecommand{\BIBforeignlanguage}[2]{{%
\expandafter\ifx\csname l@#1\endcsname\relax
\typeout{** WARNING: IEEEtran.bst: No hyphenation pattern has been}%
\typeout{** loaded for the language `#1'. Using the pattern for}%
\typeout{** the default language instead.}%
\else
\language=\csname l@#1\endcsname
\fi
#2}}
\providecommand{\BIBdecl}{\relax}
\BIBdecl

\bibitem{Dargan2019ASO}
S.~Dargan, M.~Kumar, M.~R. Ayyagari, and G.~Kumar, ``{A Survey of Deep Learning and Its Applications: A New Paradigm to Machine Learning},'' \emph{Archives of Computational Methods in Engineering}, vol.~27, pp. 1071--1092, 2020.

\bibitem{Firouzi2020IIoT}
F.~Firouzi, K.~Chakrabarty, and S.~Nassif, \emph{{Intelligent Internet of Things: From Device to Fog and Cloud}}.\hskip 1em plus 0.5em minus 0.4em\relax Springer Nature, 2020.

\bibitem{Ching2017OpportunitiesAO}
T.~Ching, D.~S. Himmelstein, B.~K. Beaulieu-Jones, A.~A. Kalinin, B.~T. Do, G.~P. Way, E.~Ferrero, P.-M. Agapow, M.~Zietz, M.~M. Hoffman, W.~Xie, G.~L. Rosen, B.~J. Lengerich, J.~Israeli, J.~Lanchantin, S.~Woloszynek, A.~E. Carpenter, A.~Shrikumar, J.~Xu, E.~M. Cofer, C.~A. Lavender, S.~C. Turaga, A.~M. Alexandari, Z.~Lu, D.~J. Harris, D.~DeCaprio, Y.~Qi, A.~Kundaje, Y.~Peng, L.~K. Wiley, M.~H.~S. Segler, S.~M. Boca, S.~J. Swamidass, A.~Huang, A.~Gitter, and C.~S. Greene, ``{Opportunities and obstacles for deep learning in biology and medicine},'' \emph{Journal of the Royal Society Interface}, vol.~15, p. 20170387, 2018.

\bibitem{Xu2021ChatbotFH}
L.~Xu, L.~Sanders, K.~Li, and J.~Chow, ``{Chatbot for Health Care and Oncology Applications Using Artificial Intelligence and Machine Learning: Systematic Review},'' \emph{JMIR Cancer}, vol.~7, no.~4, 2021.

\bibitem{Zhang2022ShiftingML}
A.~Zhang, L.~Xing, J.~Zou, and J.~C. Wu, ``{Shifting machine learning for healthcare from development to deployment and from models to data},'' \emph{Nature Biomedical Engineering}, vol.~6, pp. 1330--1345, 2022.

\bibitem{Seh2020HealthcareDB}
A.~H. Seh, M.~Zarour, M.~Alenezi, A.~K. Sarkar, A.~Agrawal, R.~Kumar, and R.~A. Khan, ``{Healthcare Data Breaches: Insights and Implications},'' \emph{Healthcare}, vol.~8, p. 133, 2020.

\bibitem{Zarour2021EnsuringDI}
M.~Zarour, M.~Alenezi, T.~J. Ansari, A.~K. Pandey, M.~Ahmad, A.~Agrawal, R.~Kumar, and R.~A. Khan, ``{Ensuring data integrity of healthcare information in the era of digital health},'' \emph{Healthcare Technology Letters}, vol.~8, pp. 66--77, 2021.

\bibitem{Neprash2022TrendsIR}
H.~T. Neprash, C.~C. McGlave, D.~A. Cross, B.~A. Virnig, M.~A. Puskarich, J.~D. Huling, A.~Z. Rozenshtein, and S.~S. Nikpay, ``{Trends in Ransomware Attacks on US Hospitals, Clinics, and Other Health Care Delivery Organizations, 2016-2021},'' \emph{JAMA Health Forum}, vol.~3, no.~12, p. e224873, 2022.

\bibitem{Murdoch2021PrivacyAA}
B.~Murdoch, ``{Privacy and artificial intelligence: challenges for protecting health information in a new era},'' \emph{BMC Medical Ethics}, vol.~22, p. 122, 2021.

\bibitem{Finlayson2019AdversarialAO}
S.~G. Finlayson, J.~Bowers, J.~Ito, J.~Zittrain, A.~Beam, and I.~S. Kohane, ``{Adversarial attacks on medical machine learning},'' \emph{Science}, vol. 363, pp. 1287--1289, 2019.

\bibitem{Argaw2020CybersecurityOH}
S.~T. Argaw, J.~R. Troncoso-Pastoriza, D.~Lacey, M.-V. Florin, F.~Calcavecchia, D.~Anderson, W.~Burleson, J.-M. Vogel, C.~O’Leary, B.~Eshaya-Chauvin, and A.~Flahault, ``{Cybersecurity of Hospitals: discussing the challenges and working towards mitigating the risks},'' \emph{BMC Medical Informatics and Decision Making}, vol.~20, p. 146, 2020.

\bibitem{Rieke2020TheFO}
N.~Rieke, J.~Hancox, W.~Li, F.~Milletar{\`i}, H.~R. Roth, S.~Albarqouni, S.~Bakas, M.~Galtier, B.~A. Landman, K.~H. Maier-Hein, S.~Ourselin, M.~J. Sheller, R.~M. Summers, A.~Trask, D.~Xu, M.~Baust, and M.~J. Cardoso, ``{The future of digital health with federated learning},'' \emph{NPJ Digital Medicine}, vol.~3, p. 119, 2020.

\bibitem{Li2019ASO}
Q.~Li, Z.~Wen, Z.~Wu, and B.~He, ``{A Survey on Federated Learning Systems: Vision, Hype and Reality for Data Privacy and Protection},'' \emph{IEEE Transactions on Knowledge and Data Engineering}, vol.~35, no.~4, pp. 3347--3366, 2019.

\bibitem{Usynin2021AdversarialIA}
D.~Usynin, A.~Ziller, M.~R. Makowski, R.~F. Braren, D.~Rueckert, B.~Glocker, G.~Kaissis, and J.~Passerat-Palmbach, ``{Adversarial interference and its mitigations in privacy-preserving collaborative machine learning},'' \emph{Nature Machine Intelligence}, vol.~3, pp. 749--758, 2021.

\bibitem{Liu2022ThreatsAA}
P.~Liu, X.~Xu, and W.~Wang, ``{Threats, attacks and defenses to federated learning: issues, taxonomy and perspectives},'' \emph{Cybersecurity}, vol.~5, p.~4, 2022.

\bibitem{He2021AttackingAP}
Z.~He, T.~Zhang, and R.~B. Lee, ``{Attacking and Protecting Data Privacy in Edge–Cloud Collaborative Inference Systems},'' \emph{IEEE Internet of Things Journal}, vol.~8, pp. 9706--9716, 2021.

\bibitem{Yang2024PracticalFI}
R.~Yang, J.~Ma, J.~Zhang, S.~Kumari, S.~Kumar, and J.~J. P.~C. Rodrigues, ``{Practical Feature Inference Attack in Vertical Federated Learning During Prediction in Artificial Internet of Things},'' \emph{IEEE Internet of Things Journal}, vol.~11, pp. 5--16, 2024.

\bibitem{Boulemtafes2020ARO}
\BIBentryALTinterwordspacing
A.~Boulemtafes, A.~Derhab, and Y.~Challal, ``{A review of privacy-preserving techniques for deep learning},'' \emph{Neurocomputing}, vol. 384, pp. 21--45, 2020. [Online]. Available: \url{https://api.semanticscholar.org/CorpusID:211159333}
\BIBentrySTDinterwordspacing

\bibitem{Ma2019UnderstandingAA}
X.~Ma, Y.~Niu, L.~Gu, Y.~Wang, Y.~Zhao, J.~Bailey, and F.~Lu, ``{Understanding Adversarial Attacks on Deep Learning Based Medical Image Analysis Systems},'' \emph{Pattern Recognition}, vol. 110, p. 107332, 2021.

\bibitem{Kaviani2022AdversarialAA}
S.~Kaviani, K.~J. Han, and I.~Sohn, ``{Adversarial attacks and defenses on AI in medical imaging informatics: A survey},'' \emph{Expert Syst. Appl.}, vol. 198, p. 116815, 2022.

\bibitem{AllenZhu2018ACT}
Z.~Allen-Zhu, Y.~Li, and Z.~Song, ``{A Convergence Theory for Deep Learning via Over-Parameterization},'' in \emph{Proceedings of the 36th International Conference on Machine Learning}, vol.~97, 2019, pp. 242--252.

\bibitem{Puttagunta2023AdversarialEA}
M.~K. Puttagunta, R.~Subban, and C.~N.~K. Babu, ``{Adversarial examples: attacks and defences on medical deep learning systems},'' \emph{Multimedia Tools and Applications}, vol.~82, pp. 33\,773--33\,809, 2023.

\bibitem{Dibbo2023SoKMI}
S.~V. Dibbo, ``{SoK: Model Inversion Attack Landscape: Taxonomy, Challenges, and Future Roadmap},'' \emph{2023 IEEE 36th Computer Security Foundations Symposium (CSF)}, pp. 439--456, 2023.

\bibitem{Fang2024PrivacyLO}
H.~Fang, Y.~Qiu, H.~Yu, W.~Yu, J.~Kong, B.~Chong, B.~Chen, X.~Wang, and S.-T. Xia, ``{Privacy Leakage on DNNs: A Survey of Model Inversion Attacks and Defenses},'' \emph{ArXiv}, vol. abs/2402.04013, 2024.

\bibitem{Nasr2018ComprehensivePA}
M.~Nasr, R.~Shokri, and A.~Houmansadr, ``{Comprehensive Privacy Analysis of Deep Learning: Passive and Active White-box Inference Attacks against Centralized and Federated Learning},'' \emph{2019 IEEE Symposium on Security and Privacy (SP)}, pp. 739--753, 2018.

\bibitem{Melis2018ExploitingUF}
L.~Melis, C.~Song, E.~D. Cristofaro, and V.~Shmatikov, ``{Exploiting Unintended Feature Leakage in Collaborative Learning},'' \emph{2019 IEEE Symposium on Security and Privacy (SP)}, pp. 691--706, 2018.

\bibitem{Wang2019EavesdropTC}
L.~Wang, S.~Xu, X.~Wang, and Q.~Zhu, ``{Eavesdrop the Composition Proportion of Training Labels in Federated Learning},'' \emph{ArXiv}, vol. abs/1910.06044, 2019.

\bibitem{Wang2023PoisoningAssistedPI}
Z.~Wang, Y.~Huang, M.~Song, L.~Wu, F.~Xue, and K.~Ren, ``{Poisoning-Assisted Property Inference Attack Against Federated Learning},'' \emph{IEEE Transactions on Dependable and Secure Computing}, vol.~20, pp. 3328--3340, 2023.

\bibitem{Zhang2019TheSR}
Y.~Zhang, R.~Jia, H.~Pei, W.~Wang, B.~Li, and D.~X. Song, ``{The Secret Revealer: Generative Model-Inversion Attacks Against Deep Neural Networks},'' \emph{2020 IEEE/CVF Conference on Computer Vision and Pattern Recognition (CVPR)}, pp. 250--258, 2019.

\bibitem{Chen2020KnowledgeEnrichedDM}
S.~Chen, M.~Kahla, R.~Jia, and G.-J. Qi, ``{Knowledge-Enriched Distributional Model Inversion Attacks},'' \emph{2021 IEEE/CVF International Conference on Computer Vision (ICCV)}, pp. 16\,158--16\,167, 2020.

\bibitem{An.Mirror.NDSS.2022}
S.~An, G.~Tao, Q.~Xu, Y.~Liu, G.~Shen, Y.~Yao, J.~Xu, and X.~Zhang, ``{MIRROR: Model Inversion for Deep Learning Network with High Fidelity},'' in \emph{Proceedings of the Network and Distributed Systems Security Symposium (NDSS 2022)}, 2022.

\bibitem{Nguyen2023ReThinkingMI}
N.-B. Nguyen, K.~Chandrasegaran, M.~Abdollahzadeh, and N.-M. Cheung, ``{Re-Thinking Model Inversion Attacks Against Deep Neural Networks},'' \emph{2023 IEEE/CVF Conference on Computer Vision and Pattern Recognition (CVPR)}, pp. 16\,384--16\,393, 2023.

\bibitem{Zhu2019DeepLF}
L.~Zhu, Z.~Liu, and S.~Han, ``{Deep Leakage from Gradients},'' in \emph{33rd Conference on Neural Information Processing Systems (NeurIPS 2019)}, 2019.

\bibitem{Geiping2020InvertingG}
J.~Geiping, H.~Bauermeister, H.~Dr{\"o}ge, and M.~Moeller, ``{Inverting Gradients - How easy is it to break privacy in federated learning?}'' in \emph{Neural Information Processing Systems}, 2020.

\bibitem{Wei2020AFF}
W.~Wei, L.~Liu, M.~L. Loper, K.-H. Chow, M.~E. Gursoy, S.~Truex, and Y.~Wu, ``{A Framework for Evaluating Gradient Leakage Attacks in Federated Learning},'' \emph{ArXiv}, vol. abs/2004.10397, 2020.

\bibitem{Yin2021SeeTG}
H.~Yin, A.~Mallya, A.~Vahdat, J.~M. {\'A}lvarez, J.~Kautz, and P.~Molchanov, ``{See through Gradients: Image Batch Recovery via GradInversion},'' \emph{2021 IEEE/CVF Conference on Computer Vision and Pattern Recognition (CVPR)}, pp. 16\,332--16\,341, 2021.

\bibitem{Wu2022LearningTI}
R.~Wu, X.~Chen, C.~Guo, and K.~Q. Weinberger, ``{Learning to invert: simple adaptive attacks for gradient inversion in federated learning},'' in \emph{Proceedings of the Thirty-Ninth Conference on Uncertainty in Artificial Intelligence}, 2023, p. 214.

\bibitem{Geng2023ImprovedGI}
J.~Geng, Y.~Mou, Q.~Li, F.~Li, O.~D. Beyan, S.~Decker, and C.~Rong, ``{Improved Gradient Inversion Attacks and Defenses in Federated Learning},'' \emph{IEEE Transactions on Big Data}, 2023, early Access.

\bibitem{Li2022AuditingPD}
Z.~Li, J.~Zhang, L.~Liu, and J.~Liu, ``{Auditing Privacy Defenses in Federated Learning via Generative Gradient Leakage},'' \emph{2022 IEEE/CVF Conference on Computer Vision and Pattern Recognition (CVPR)}, pp. 10\,122--10\,132, 2022.

\bibitem{Jia2018AttriGuardAP}
J.~Jia and N.~Z. Gong, ``{AttriGuard: A Practical Defense Against Attribute Inference Attacks via Adversarial Machine Learning},'' in \emph{USENIX Security Symposium}, 2018.

\bibitem{Mireshghallah2019ShredderLN}
F.~Mireshghallah, M.~Taram, P.~Ramrakhyani, D.~M. Tullsen, and H.~Esmaeilzadeh, ``{Shredder: Learning Noise to Protect Privacy with Partial DNN Inference on the Edge},'' \emph{ArXiv}, vol. abs/1905.11814, 2019.

\bibitem{Sun2020SoteriaPD}
J.~Sun, A.~Li, B.~Wang, H.~Yang, H.~Li, and Y.~Chen, ``{Soteria: Provable Defense against Privacy Leakage in Federated Learning from Representation Perspective},'' \emph{2021 IEEE/CVF Conference on Computer Vision and Pattern Recognition (CVPR)}, pp. 9307--9315, 2020.

\bibitem{Mireshghallah2021NotAF}
F.~Mireshghallah, M.~Taram, A.~Jalali, A.~T.~T. Elthakeb, D.~Tullsen, and H.~Esmaeilzadeh, ``{Not All Features Are Equal: Discovering Essential Features for Preserving Prediction Privacy},'' in \emph{Proceedings of the Web Conference 2021}.\hskip 1em plus 0.5em minus 0.4em\relax Association for Computing Machinery, 2021, pp. 669--680.

\bibitem{Zhu2022AFD}
L.~Zhu, X.~Liu, Y.~Li, X.~Yang, S.~Xia, and R.~Lu, ``{A Fine-Grained Differentially Private Federated Learning Against Leakage From Gradients},'' \emph{IEEE Internet of Things Journal}, vol.~9, pp. 11\,500--11\,512, 2022.

\bibitem{Shen2022PerformanceEnhancedFL}
X.~Shen, Y.~Liu, and Z.~Zhang, ``{Performance-Enhanced Federated Learning With Differential Privacy for Internet of Things},'' \emph{IEEE Internet of Things Journal}, vol.~9, pp. 24\,079--24\,094, 2022.

\bibitem{Cui2023RecUPFLRU}
Y.~Cui, S.~I.~A. Meerza, Z.~Li, L.~Liu, J.~Zhang, and J.~Liu, ``{RecUP-FL: Reconciling Utility and Privacy in Federated learning via User-configurable Privacy Defense},'' in \emph{Proceedings of the 2023 ACM Asia Conference on Computer and Communications Security}.\hskip 1em plus 0.5em minus 0.4em\relax Association for Computing Machinery, 2023, pp. 80--94.

\bibitem{Roy2019MitigatingIL}
P.~C. Roy and V.~N. Boddeti, ``{Mitigating Information Leakage in Image Representations: A Maximum Entropy Approach},'' \emph{2019 IEEE/CVF Conference on Computer Vision and Pattern Recognition (CVPR)}, pp. 2581--2589, 2019.

\bibitem{Wang2020ImprovingRT}
T.~Wang, Y.~Zhang, and R.~Jia, ``{Improving Robustness to Model Inversion Attacks via Mutual Information Regularization},'' in \emph{The Thirty-Fifth AAAI Conference on Artificial Intelligence}, 2021.

\bibitem{Peng2022BilateralDO}
X.~Peng, F.~Liu, J.~Zhang, L.~Lan, J.~Ye, T.~Liu, and B.~Han, ``{Bilateral Dependency Optimization: Defending Against Model-inversion Attacks},'' in \emph{Proceedings of the 28th ACM SIGKDD Conference on Knowledge Discovery and Data Mining}.\hskip 1em plus 0.5em minus 0.4em\relax Association for Computing Machinery, 2022, pp. 1358--1367.

\bibitem{Hamm2016MinimaxFL}
J.~Hamm, ``{Minimax Filter: Learning to Preserve Privacy from Inference Attacks},'' \emph{J. Mach. Learn. Res.}, vol.~18, pp. 129:1--129:31, 2016.

\bibitem{Stock2023LessonsLD}
J.~Stock, J.~Wettlaufer, D.~Demmler, and H.~Federrath, ``{Lessons Learned: Defending Against Property Inference Attacks},'' in \emph{International Conference on Security and Cryptography}, 2023.

\bibitem{Gong2023AGD}
X.~Gong, Z.~Wang, S.~Li, Y.~Chen, and Q.~Wang, ``{A GAN-Based Defense Framework Against Model Inversion Attacks},'' \emph{IEEE Transactions on Information Forensics and Security}, vol.~18, pp. 4475--4487, 2023.

\bibitem{Sandeepa2024RecDefAR}
C.~Sandeepa, B.~Siniarski, S.~Wang, and M.~Liyanage, ``{Rec-Def: A Recommendation-Based Defence Mechanism for Privacy Preservation in Federated Learning Systems},'' \emph{IEEE Transactions on Consumer Electronics}, vol.~70, pp. 2716--2728, 2024.

\bibitem{Chen2022FederatedLA}
Y.~Chen, Y.~Gui, H.~Lin, W.~Gan, and Y.~Wu, ``{Federated Learning Attacks and Defenses: A Survey},'' \emph{2022 IEEE International Conference on Big Data (Big Data)}, pp. 4256--4265, 2022.

\bibitem{Li2024ThreatsAD}
Y.~Li, Z.~Guo, N.~Yang, H.~Chen, D.~Yuan, and W.~Ding, ``{Threats and Defenses in Federated Learning Life Cycle: A Comprehensive Survey and Challenges},'' \emph{ArXiv}, vol. abs/2407.06754, 2024.

\bibitem{Kermani2015SystematicPA}
M.~Mozaffari-Kermani, S.~Sur-Kolay, A.~Raghunathan, and N.~K. Jha, ``{Systematic Poisoning Attacks on and Defenses for Machine Learning in Healthcare},'' \emph{IEEE Journal of Biomedical and Health Informatics}, vol.~19, no.~6, pp. 1893--1905, 2015.

\bibitem{Newaz2020AdversarialAT}
A.~I. Newaz, N.~I. Haque, A.~K. Sikder, M.~A. Rahman, and A.~S. Uluagac, ``{Adversarial Attacks to Machine Learning-Based Smart Healthcare Systems},'' \emph{GLOBECOM 2020 - 2020 IEEE Global Communications Conference}, pp. 1--6, 2020.

\bibitem{Rahman2020AdversarialET}
A.~Rahman, M.~S. Hossain, N.~A. Alrajeh, and F.~J. Alsolami, ``{Adversarial Examples—Security Threats to COVID-19 Deep Learning Systems in Medical IoT Devices},'' \emph{IEEE Internet of Things Journal}, vol.~8, pp. 9603--9610, 2020.

\bibitem{Grama2020RobustAF}
M.~Grama, M.~Mușat, L.~Mu{\~n}oz-Gonz{\'a}lez, J.~Passerat-Palmbach, D.~Rueckert, and A.~Alansary, ``{Robust Aggregation for Adaptive Privacy Preserving Federated Learning in Healthcare},'' \emph{ArXiv}, vol. abs/2009.08294, 2020.

\bibitem{Lin2023PrivacyAwareAC}
H.~Lin, K.~Kaur, X.~Wang, G.~Kaddoum, J.~Hu, and M.~M. Hassan, ``{Privacy-Aware Access Control in IoT-Enabled Healthcare: A Federated Deep Learning Approach},'' \emph{IEEE Internet of Things Journal}, vol.~10, pp. 2893--2902, 2023.

\bibitem{Ali2022FederatedLF}
M.~Ali, F.~Naeem, M.~A. Tariq, and G.~Kaddoum, ``{Federated Learning for Privacy Preservation in Smart Healthcare Systems: A Comprehensive Survey},'' \emph{IEEE Journal of Biomedical and Health Informatics}, vol.~27, pp. 778--789, 2022.

\bibitem{Muoka2023ACR}
G.~W. Muoka, D.~Yi, C.~C. Ukwuoma, A.~Mutale, C.~J. Ejiyi, A.~K. Mzee, E.~S.~A. Gyarteng, A.~Alqahtani, and M.~A. Al-antari, ``{A Comprehensive Review and Analysis of Deep Learning-Based Medical Image Adversarial Attack and Defense},'' \emph{Mathematics}, vol.~11, no.~20, 2023.

\bibitem{Benesty2009PearsonCC}
J.~Benesty, J.~Chen, Y.~Huang, and I.~Cohen, \emph{{Pearson Correlation Coefficient}}.\hskip 1em plus 0.5em minus 0.4em\relax Springer Berlin Heidelberg, 2009, pp. 1--4.

\bibitem{Sun2023GeneratingSP}
C.~Sun, J.~van Soest, and M.~Dumontier, ``{Generating synthetic personal health data using conditional generative adversarial networks combining with differential privacy},'' \emph{Journal of biomedical informatics}, vol. 143, p. 104404, 2023.

\bibitem{Ho2020TheRC}
Y.~Ho and S.~Wookey, ``{The Real-World-Weight Cross-Entropy Loss Function: Modeling the Costs of Mislabeling},'' \emph{IEEE Access}, vol.~8, pp. 4806--4813, 2020.

\bibitem{Krizhevsky2009LearningML}
A.~Krizhevsky, ``Learning multiple layers of features from tiny images,'' University of Toronto, Tech. Rep., 2009.

\bibitem{medmnistv2}
J.~Yang, R.~Shi, D.~Wei, Z.~Liu, L.~Zhao, B.~Ke, H.~Pfister, and B.~Ni, ``{MedMNIST v2-A large-scale lightweight benchmark for 2D and 3D biomedical image classification},'' \emph{Scientific Data}, vol.~10, no.~1, p.~41, 2023.

\bibitem{Abadi2016DeepLW}
M.~Abadi, A.~Chu, I.~Goodfellow, H.~B. McMahan, I.~Mironov, K.~Talwar, and L.~Zhang, ``{Deep Learning with Differential Privacy},'' in \emph{Proceedings of the 2016 ACM SIGSAC Conference on Computer and Communications Security}.\hskip 1em plus 0.5em minus 0.4em\relax Association for Computing Machinery, 2016, pp. 308--318.

\bibitem{He2015DeepRL}
K.~He, X.~Zhang, S.~Ren, and J.~Sun, ``{Deep Residual Learning for Image Recognition},'' \emph{2016 IEEE Conference on Computer Vision and Pattern Recognition (CVPR)}, pp. 770--778, 2015.

\bibitem{scikit-learn}
F.~Pedregosa, G.~Varoquaux, A.~Gramfort, V.~Michel, B.~Thirion, O.~Grisel \emph{et~al.}, ``{Scikit-learn: Machine Learning in {P}ython},'' \emph{Journal of Machine Learning Research}, vol.~12, pp. 2825--2830, 2011.

\bibitem{NEURIPS2019_9015}
A.~Paszke, S.~Gross, F.~Massa, A.~Lerer, J.~Bradbury, G.~Chanan \emph{et~al.}, ``{PyTorch: An Imperative Style, High-Performance Deep Learning Library},'' in \emph{Advances in Neural Information Processing Systems 32}, H.~Wallach, H.~Larochelle, A.~Beygelzimer, F.~d\textquotesingle Alch\'{e}-Buc, E.~Fox, and R.~Garnett, Eds.\hskip 1em plus 0.5em minus 0.4em\relax Curran Associates, Inc., 2019, pp. 8024--8035.

\bibitem{He2014ConvolutionalNN}
K.~He and J.~Sun, ``{Convolutional neural networks at constrained time cost},'' \emph{2015 IEEE Conference on Computer Vision and Pattern Recognition (CVPR)}, pp. 5353--5360, 2014.

\bibitem{Lyu2020ThreatsTF}
L.~Lyu, H.~Yu, and Q.~Yang, ``{Threats to Federated Learning: A Survey},'' \emph{ArXiv}, vol. abs/2003.02133, 2020.

\bibitem{Yang2024ALA}
Y.~Yang, Q.~Li, C.~Nie, Y.~Hong, M.~Pang, and B.~Wang, ``{A Learning-Based Attack Framework to Break SOTA Poisoning Defenses in Federated Learning},'' \emph{ArXiv}, vol. abs/2407.15267, 2024.

\bibitem{Huang2021EvaluatingGI}
Y.~Huang, S.~Gupta, Z.~Song, K.~Li, and S.~Arora, ``{Evaluating Gradient Inversion Attacks and Defenses in Federated Learning},'' in \emph{35th Conference on Neural Information Processing Systems (NeurIPS 2021)}, 2021.

\end{thebibliography}

\end{document}